\pdfoutput=1

\documentclass[11pt]{article}

\usepackage[]{EMNLP2023}

\usepackage{times}
\usepackage{latexsym}
\usepackage{graphicx} 
\usepackage[T1]{fontenc}
\usepackage{multirow}
\usepackage[utf8]{inputenc}
\usepackage{microtype}

\usepackage{inconsolata}

\title{Inference is All You Need:\\ Self Example Retriever for Cross-domain Dialogue State Tracking with ChatGPT}

\author{Jihyun Lee$^1$, Gary Geunbae Lee$^{1,2}$ \\
  $^1$Graduate School of Artificial Intelligence, POSTECH, Republic of Korea\\
  $^2$Department of Computer Science and Engineering, POSTECH, Republic of Korea\\
  \texttt{\{jihyunlee,  gblee\}@postech.ac.kr} \\
}

\begin{document}
\maketitle
\begin{abstract}
Traditional dialogue state tracking approaches heavily rely on extensive training data and handcrafted features, limiting their scalability and adaptability to new domains. In this paper, we propose a novel method that leverages inference and in-context learning with ChatGPT for domain transfer in dialogue state tracking, without any parameter updates. By guiding ChatGPT's chain of thought, we enable it to retrieve relevant examples and generalize knowledge to accurately infer dialogue states, solely through inference. Experimental results on the MultiWOZ dataset demonstrate competitive performance and promising generalization across domains. Our parameter-free approach offers a scalable and adaptable solution, opening new research directions in domain transfer learning.
\end{abstract}

\section{Introduction}
\begin{figure*}[h]
    \centering
    \includegraphics[width=450pt]{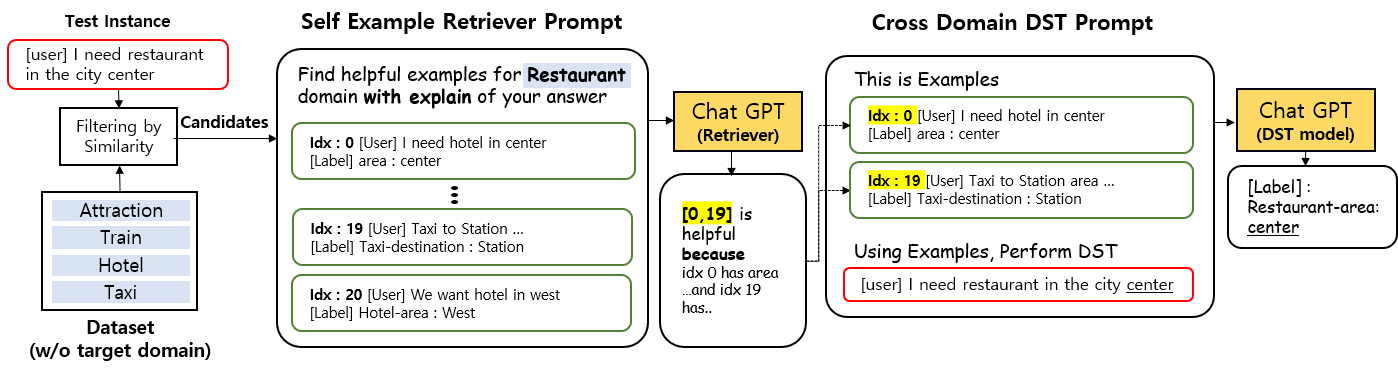}
    \caption{Overview of SERI-DST.}
    \label{fig:main}
\end{figure*}

Dialogue State Tracking (DST) is a crucial component of task-oriented dialogue systems, predicting essential conversational information \cite{young2010hidden}. A common limitation in current DST systems is the requirement to pre-define the domain before the training. While this is manageable in research, it poses challenges in practical applications where users frequently request new domains.
For example, consider a DST system designed to provide bus and airplane information. Users may also want to get the train information from the same service. In conventional DST, integrating a new domain involves the cumbersome process of collecting and annotating domain-specific data, followed by retraining the entire model. This inflexibility in accommodating new domains highlights the need for more versatile and efficient approaches. \footnote{In the field of DST, the term "Zero-shot DST" (held-out target domain when training) is often used to refer to Cross-domain DST \cite{trade, lin2021zero}. To provide clarity, we specifically use the term "Cross-domain" instead of "Zero-shot."}.

Several strategies have been proposed for achieving domain transfer to unseen domains by utilizing natural language form slot descriptions  \cite{lin2021leveraging, heck2022robust}, leveraging question answering dataset \cite{li2021zero, sf-dst}, and augmentation with ontology \cite{campagna2020zero}. However, existing methods encounter two key challenges in terms of explainability. Firstly, it is often unclear which existing domains have contributed to the successful transition to unseen domain. Secondly, it's uncertain which characteristics of the dataset drive effective domain adaptation — whether shorter, concise utterances or information-rich, longer ones are more advantageous. Understanding the model's behavior is essential for strategic planning in collecting additional data for further training and making the model adapt to a new domain. Unfortunately, current methods operate as black boxes, limiting the utility of model insights for such planning.

To this end, we developed an explainable example retrieval and In-Context Learning system for the Cross-Domain Dialogue State Tracking (DST) model by leveraging the LLM. Recent research has shown that Language Models (LLMs) can generate reliable natural language explanations. Building upon this work, we utilized LLM to create transparent and interpretable explanations for understanding the model's behavior and planning future data collection strategies. Here, we query to the LLM by asking, \textit{"Is this example useful for solving this test? why?"}. and get the answers and explanation of why it is helpful.

Furthermore, by utilizing the In-Context Learning approach, we effectively employ the retrieved results to generate DST result for unseen domain. In-context learning allows for a more transparent inspection of the inference process for new domains compared to traditional parameter update methods. This characteristic of the in-context learning approach makes it easy to understand the model's inference steps, which is well-suited to the challenges encountered in cross-domain DST research. Here, to enhance the coherance, we use same model for retreival and DST.

\section{Problem Statement}
\textbf{Dialogue state tracking }. In DST task, a dialogue set, denoted as $D$, consists of individual dialogues represented as $\left\{d_n\right\}_{n=1}^{N}$ for $N$ number of dialogue. Each dialogue $d_n$ comprises a sequence of paired user and system utterances $\left\{(u_t^n,s_t^n) \right\}_{t=1}^{T}$  and turn-level belief states $b^n_t$ at each turn $t$. A turn-level belief states $b^n_t$ includes slot-value pairs, where a slot represents a specific category of information (e.g., hotel-area) and the value is the corresponding information (e.g., west). The accumulated belief state up to turn $t$ is represented as $B^n_t = \{b^n_1, b^n_2, ..., b^n_t\}$. For clarity, we denote the dialogue set $D$ excluding the target domain as $D^{-target}$.
\\

\noindent
\textbf{Example retriever for domain transfer}. In in-context learning, the provided example within the prompt is critical for achieving reliable downstream task performance \cite{liu2021makes}. This observation has sparked a research interest in the field of example retrieving tasks \cite{das2021case,rubin2021learning,cheng2023uprise}. Particularly in a domain transfer scenario, given the dataset $D^{-target}$, the example retriever finds useful examples that can aid in efficiently performing in the target domain.
\\

\section{Method}
\subsection{Self Example Retriever}

To get the relevant example for the downstream task without training a specific retriever, we leveraged the LLM itself as a retriever. To reduce the computational resources for retrieving, we first filtered $K$ candidate examples from the $D^{-target}$ dataset, using the text similarity \cite{bm25} by giving test instance $u_{test}$. Then, we concatenate the filtered candidates $\left\{(u_i, b_i)\right\}_{i=0}^K$ with their index $i$ and query to LLM to infer $M$ number of useful example indexes among the given candidates. Additionally, we request explanations for the model's retrieved examples to facilitate complex reasoning tasks \cite{cot}.  This prompt allows us to extract accurate and relevant examples from $D^{-target}$. The full prompt can be found in Appendix~\ref{sec:prompt1}, and we set $K$ as 20, considering the maximum input length of the LLM.

\subsection{Cross-domain DST}
To perform the in-context learning with retrieved examples, we construct the prompt by concatenating the retrieved examples. Additionally, for helping transfer the knowledge, we add the slot name and its detailed explanation to give a hint for the unseen domain, followed by the previous researches \cite{heck2020trippy, hu2022context, heck2023chatgpt}. Lastly, we concat the test instance $u_{test}$ with the latest history ($u_{test-1}, s_{test-1}$) to get the result. From this prompt, we obtain the predicted $b_{test}$ as an output (Figure~\ref{fig:main}). The full prompt is in Appendix~\ref{sec:prompt2}.
\section{Experiments}
\subsection{Dataset and Metric} In order to assess the effectiveness of our SERI-DST framework, we perform experiments using the MultiWOZ2.1 dataset \cite{woz}. This dataset comprises 10,000 multi-turn dialogues that span five different domains\footnote{Attraction, Hotel, Restaurant, Taxi, and Train}.
To evaluate the performance, we utilize the domain joint goal accuracy (DomainJGA) metric. DomainJGA measures the ratio of correctly predicted turns to the total number of turns with respect to a target domain \cite{trade}. A turn is marked correct if all of its predicted target domain slot-value pairs in $B_t$ match the ground truth.
\subsection{Comparison with Baselines}
\begin{table}[h]

\centering
\scalebox{0.68}{
\begin{tabular}{l|lllll|l}
\hline
Models     & attr. & hotel & rest. & taxi & train & avg   \\\hline
TRADE (\citeyear{wu2019transferable})      & 22.8  & 19.5  & 16.4  & 59.2 & 22.9  & 28.16 \\
TripPy-R (\citeyear{heck2020trippy})  & 27.1  & 18.3  & 15.3  & 61.5 & 23.7  & 29.18 \\
TransferQA (\citeyear{lin2021zero}) & 31.3  & 22.7  & 26.3  & 61.9 & 36.7  & 35.78 \\
\citet{li2021zero}   & 42.4  & 24.9  & 27.7  & 60.3 & 41.1  & 39.28 \\
D3ST (\citeyear{zhao2022description})      & 56.4  & 21.8  & 38.2  & \textbf{78.4} & 38.7  & 46.70 \\
\citet{campagna2020zero}     & 52.8  & 36.3  & 45.3  & 62.6 & 46.7  & 48.74 \\
ChatGPT-DST (\citeyear{heck2023chatgpt})    & 52.7  & 42.0  & \textbf{55.8}  & 70.9 & \textbf{60.8}  & 56.44 \\\hline
SERI-DST   & \textbf{67.5}  & \textbf{47.0}  & 51.7  & 77.1 & 59.5  & \textbf{60.58} \\\hline\hline
IC-DST (\citeyear{hu2022context})     & 60.0  & 46.7  & 57.3  & 71.4 & 49.4  & 56.96 \\ \hline
\end{tabular}
}
\caption{Comparison DomainJGA with other DST models performance on MultiWOZ dataset.}
\label{tab:main}
\end{table}

In this experiment, we benchmarked the SERI-DST model against other state-of-the-art models. As demonstrated in Table ~\ref{tab:main}, SERI-DST showed the highest average score and excelled in two out of five domains. Two key insights emerge from these findings: First, the ChatGPT-DST \cite{heck2023chatgpt}, a recently employed ChatGPT \cite{ChatGPT} for DST, performs prediction using only a slot description and dialogue history without any examples. Our model's superior performance suggests that providing self-retrieved examples enhanced DST prediction.
Second, our method outperforms the IC-DST model, which uses an in-context method like ours. However, while IC-DST needs to train additional retriever model to get good examples, our model shows better results without a specifically trained retriever by leveraging LLM.

\subsection{Ablation Study}

\begin{table}[h]

\centering
\scalebox{0.73}{
\begin{tabular}{l|lllll|c}
\hline

Ret. Method            & attr. & hotel & rest. & taxi & train & avg\\\hline
Random\_3  & 63.9 & 28.0 & 51.3 & 63.5 & 41.7 & 50.5 \\\hline
SERI\_Top\_1   &67.2 & 44.8 & 49.9 & \textbf{78.0} & 59.6 & 59.9  \\
SERI\_Top\_2  &66.6 & 45.2 & 51.9 & 76.7& \textbf{61.5} & 60.4 \\
SERI\_Top\_3  & 67.5 & \textbf{47.0} & \textbf{51.7} & 77.1  & 59.5 & \textbf{60.6} \\\hline
\quad{w/o Explain\_3}   & \textbf{70.1} & 44.8 & 51.3 & 76.6 & 59.2  & 60.4   \\ \hline

\end{tabular}
}

\caption{Ablation study result of self example retriever. The number in each method represents the retrieved examples number.}
\label{tab:ablation}
\end{table}

To assess the impact of each component in our self-retrieval methods, we performed an ablation study by incrementally adding each component to ChatGPT \cite{ChatGPT}. As depicted in Table~\ref{tab:ablation}, the inclusion of a random example markedly low performance across the domains. Conversely, the integration of self-retrieved examples led to a performance increase, underlining the efficacy of our retriever. We also observed a positive correlation between the number of examples and performance enhancement. Additionally, we conducted the retrieval without asking for an explanation for the answer, which resulted in a minor drop in the overall score. We left this aspect as a potential focus for future research, given its promising prospects for improvement.
\section{Analysis}

\begin{table*}[h!]

\centering
\scalebox{0.7}{
\begin{tabular}{p{25mm}|p{173mm}}

\hline 
\textbf{Explain Type 1}  & \textbf{Informative Clear-Intent} \\\hline
\multirow{3}{*}{Example}       & "The user  \textcolor{blue}{specifies} the departure and destination cities and the system can  \textcolor{blue}{easily extract} the required slots.", \\  
                                & "This example is helpful because it  \textcolor{blue}{directly mentions} the desired attraction type, which is a nightclub."\\
                                & "Similar to example 2, this example  \textcolor{blue}{includes all the necessary information} and uses  \textcolor{blue}{clear language}."\\\hline
\textbf{Explain Type 2} & \textbf{Cross-Domain Correlative} \\\hline
\multirow{2}{*}{Example} 
                        & "This example provides both the departure location and time, as well as the arrival time. This is \textcolor{blue}{helpful for inferring -departure and -arrive slots.}"    \\
                         &  "It provides information about the day and number of people, which \textcolor{blue}{can be easily transferred to the taxi domain}"\\\hline
\textbf{Explain Type 3}  & \textbf{Definitive End-State} \\\hline
\multirow{2}{*}{Example}
                            & "It contains the exact phrase 'that was all I needed' which is a strong indicator that the user has \textcolor{blue}{completed their task} and \textcolor{blue}{no further information} is required."    \\
                            & "It clearly indicates the \textcolor{blue}{end of conversation} and not provide any \textcolor{blue}{further information}."\\

\hline

\end{tabular}
}
\caption{Common types of reasoning in SERI-DST example retrieval.}
\label{tab:reason}
\end{table*}

\subsection{Error Analysis}

\begin{figure}[h]
    \centering
    \includegraphics[width=215pt]{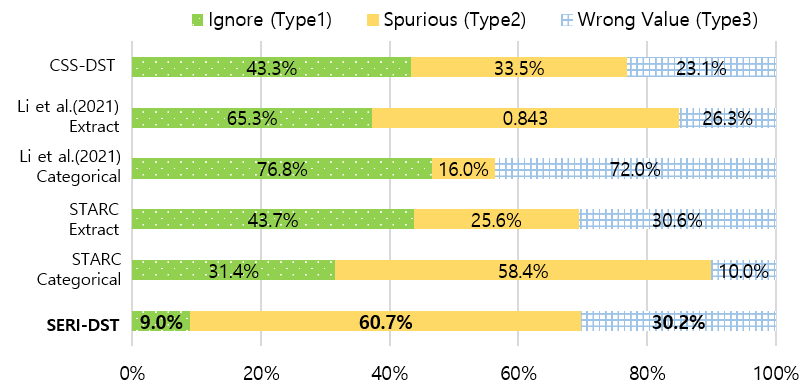}
    \caption{Error analysis of SERI-DST and other fine-tuned DST models.}
    \label{fig:error}
\end{figure}

To gain insight into the errors made by the SERI-DST model, we conducted an error analysis, comparing it with other DST models - CSS-DST \cite{zhang2022css}, \citet{li2021zero}, and STARC \cite{gao2020machine}, all specifically fine-tuned for DST. In Figure~\ref{fig:error} "Ignore" refers to when the model misses a slot, "Spurious" when it predicts a not mentioned slot, and "Wrong" when it correctly identifies a slot but gets the value wrong.
Interestingly, the LLM-based SERI-DST model showed a unique distribution of error types. Unlike other fine-tuned DST models, which distribute their errors across all categories, the SERI-DST model primarily errors in the "Spurious" category. This suggests that the model tends to provide more responses than required, even when the user does not mention them explicitly. 

Further error analysis revealed specific patterns. For example, if users only mention a restaurant's name, the LLM, having been trained on general knowledge from the web, tries to predict the type of food by the given name. Similarly, when an attraction's name is mentioned, the LLM additionally predicts its category as the answer (examples provided in Appendix~\ref{sec:error}). While this increases the "Spurious" type error, it also represents an intriguing characteristic of the LLM and highlights the potential for a more informative DST model using pre-trained knowledge.

\subsection{Exploring Reasoning }

To better understand how the LLM selects examples, we sorted the generated explanation types into three categories, shown in Table~\ref{tab:reason}. The first category is characterized by examples with clear intentions and abundant information. The LLM often considers these types of examples as beneficial for performing cross-domain DST due to their clarity. The second category involves the LLM finding connections between the example and the target domain. For instance, there are many parallels between the train and taxi domains in terms of their requirements. However, this straightforward correlation is not very common, making it less frequent than the first category. The third category involves instances where the model values utterances with clear 'endings'. These simplify the task by removing the need for tracking further information.

\subsection{Domain Influence on Transfer Learning}
\begin{figure}[h]
    \centering
    \includegraphics[width=180pt]{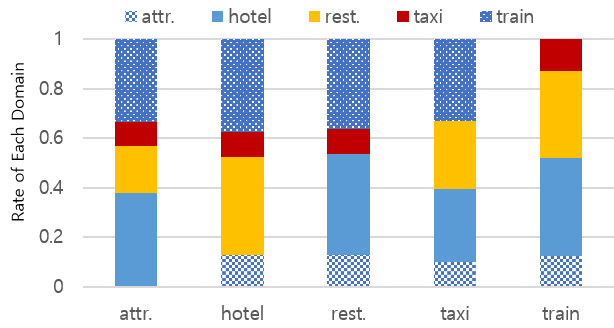}
    \caption{Impact of different domains on transfer learning.}
    \label{fig:chart}
\label{fig:transfer}
\end{figure}

In this study, we figure out which domains help transfer learning to a target domain (Figure~\ref{fig:transfer}). For this analysis, we collected the examples retrieved for the target domain and analyzed the domains contained in its label. Overall, the hotel and train domains generally help with knowledge transfer, as they tend to be frequent and informative in the MultiWOZ dataset. Interestingly, while the taxi domain usually does not contribute significantly to other domains, it had a slightly more impact on the train domain. This can be attributed to the shared attributes such as destination and departure slots between these domains. From these observations, we can affirm that the performance of the self-retrieval in the SERI-DST aligns well with human intuition.

\section{Conclusion}
We introduce a unique inference-only in-context learning framework that deploys the LLM as an integrated retriever for DST prompts, specifically designed for ChatGPT. This unique approach not only conserves computational resources but also elevates performance, which is comparable to fine-tuned retrievers and DST models. Furthermore, our analyses reveal the intriguing characteristics of leveraging the LLM for its reasoning and error cases. We anticipate a broader application for our self-retrieval method within the realm of in-context learning tasks.

\section*{Limitations}
While our research presents a promising approach to cross-domain dialogue state tracking using ChatGPT, there are several limitations that should be acknowledged:

\noindent
\textbf{Reliance on a single large language model} Our method heavily relies on the capabilities and performance of ChatGPT as the primary source of domain knowledge transfer. If ChatGPT fails to retrieve relevant examples or provides inaccurate information, it can negatively impact the performance of our approach.

\noindent
\textbf{Limited evaluation on specific domains and dataset bias} While our method demonstrates competitive performance on the MultiWOZ dataset, which encompasses a wide range of domains, it is important to note that the evaluation is primarily focused on specific domains included in the dataset. Additionally, if the MultiWOZ dataset contain bias or limitations in representing real-world dialogue scenarios, it may affect the generalization of our approach.



\bibliography{anthology,custom}
\bibliographystyle{acl_natbib}

\appendix

\section{Self Example Retriever Prompt}
\label{sec:prompt1}
\scriptsize{{\fontfamily{qcr}\selectfont
I'm finding helpful exampels to solve following dialgoue state tracking problem in domain transfer enviroment\\
curr : [user] i need a taxi at ian hong house to leave by 14:45 .\\
slots to be inference : ['-area', '-name', '-type']\\
\noindent
for attraction domain \\
please return the most useful 3 example's from below, with simple explanation why it is helpful than others for domain transfer attraction
\\\noindent
Example Number : 0\\
curr : [user] i am looking for ian hong house\\
label: None\\
\\\noindent
Example Number : 1\\
curr : [user] i need to leave after 14:45 on thursday .\\
label: train-day : thursday, train-destination : cambridge, train-leave : 14:45\\
\\\noindent
Example Number : 2\\
curr : [user] i need to arrive at the airport by 14:45 on saturday .\\
label: train-arrive : 14:45, train-day : saturday\\
\\\noindent
Example Number : 3\\
curr : [user] i need a taxi from finches bed and breakfast to ian hong house . please be here at 13:45 . what is your phone number ? what type of car ?\\
label: taxi-departure : finches bed and breakfast, taxi-leave : 13:45\\
\\\noindent
Example Number : 4\\
curr : [user] i also need a taxi from vue cinema to curry garden . i need to leave curry garden by 14:45\\
label: taxi-departure : vue cinema, taxi-destination : curry garden, taxi-leave : 14:45\\
\\\noindent
Example Number : 5\\
curr : [user] thanks ! i'd also like to book a table at ian hong house . it will need to be on the same day , same people , and we'd like to eat at 13:30 .\\
label: restaurant-day : thursday, restaurant-people : 7, restaurant-time : 13:30\\
\\\noindent
Example Number : 6\\
curr : [user] i need to leave after 14:45 .\\
label: train-leave : 14:45\\
\\\noindent
Example Number : 7\\
curr : [user] i would like the taxi to leave the restaurant by 14:45 .\\
label: taxi-departure : rice house, taxi-destination : little saint marys church, taxi-leave : 14:45\\
\\\noindent
Example Number : 8\\
curr : [user] i also need a taxi between whale of a time and lan hong house that arrives by are reservation time at the restaurant\\
label: taxi-arrive : 20:00, taxi-departure : whale of a time, taxi-destination : lan hong house\\
\\\noindent
Example Number : 9\\
curr : [user] i need to leave after 14:45 .\\
label: train-leave : 14:45\\
\\\noindent
Example Number : 10\\
curr : [user] can i have a taxi , please ? i want to leave vue cinema at 14:45 .\\
label: taxi-departure : vue cinema, taxi-leave : 14:45\\
\\\noindent
Example Number : 11\\
curr : [user] i need to arrive to ian hong house by 13:45 .\\
label: taxi-destination : ian hong\\
\\\noindent
Example Number : 12\\
curr : [user] thanks ! i am also looking for information on a restaurant named the ian hong house .\\
label: None\\
\\\noindent
Example Number : 13\\
curr : [user] thank you , can you also find a restaurant called ian hong house .\\
label: None\\
\\\noindent
Example Number : 14\\
curr : [user] i want to leave after 14:45 .\\
label: train-leave : 14:45\\
\\\noindent
Example Number : 15\\
curr : [user] yes , i need to arrive at bishops by 14:45 .\\
label: train-arrive : 14:45\\
\\\noindent
Example Number : 16\\
curr : [user] i need to arrive by 14:45 .\\
label: train-arrive : 14:45\\
\\\noindent
Example Number : 17\\
curr : [user] i would like to leave the restaurant by 14:45 .\\
label: taxi-leave : 14:45\\
\\\noindent
Example Number : 18\\
curr : [user] i want to leave at least by 14:45 if that is okay \\
label: taxi-leave : 14:45\\
\\\noindent
Example Number : 19\\
curr : [user] i need to taxi from ian hong house .\\
label: taxi-departure : lan hong house\\
\\\noindent
Output format must be '{answer : [], explanation : ), to be parsed easily.\\}
}}

\section{Cross Domain DST Prompt}
\label{sec:prompt2}
{
\scriptsize{{\fontfamily{qcr}\selectfont

> This is example of dialogue state tracking, which extract useful information from user's dailgoue. \\
> Don't guessing not mentioned information from user\\
> example's slots are \\
> hotel-area hotel-day hotel-internet hotel-name hotel-parking hotel-people hotel-pricerange hotel-stars hotel-stay hotel-type restaurant-area restaurant-day restaurant-food restaurant-name restaurant-people restaurant-pricerange restaurant-time taxi-arrive taxi-departure taxi-destination taxi-leave train-arrive train-day train-departure train-destination train-leave train-people\\
\\\noindent
------------------------------------------------------------------------------\\
\# example 1\\
dialogue: \\
prev : None\\
curr : [user] can you help me find a restaurant with a moderate price range that serves turkish food ?\\
label: restaurant-food : turkish, restaurant-pricerange : moderate\\
\\\noindent
\#example 2\\
dialogue: \\
prev : None\\
curr : [user] can you help me find a moderate restaurant that serves north american food ? thanks .\\
label: restaurant-food : north american, restaurant-pricerange : moderate\\
\\\noindent
\# example 3\\
dialogue: \\
prev : [user] i need a hotel with free wifi and free parking , thank you [system] allenbell is cheap and has free internet .\\
curr : [user] i'm looking for something in the moderate price range actually .\\
label: hotel-pricerange : moderate\\
------------------------------------------------------------------------------\\
> End of Example. In this time, the slots are extended to attraction. \\
\\\noindent
slots to be inference is\\
\\\noindent
hotel-pricerange: The price range of a hotel.\\
hotel-type: The type or category of a hotel\\
hotel-parking: Specifies whether a hotel offers parking options.\\
hotel-day: The specific day of the week for a hotel.\\
hotel-stars: The star rating of a hotel, indicating the level of quality .\\
hotel-stay: The duration or length of stay in a hotel.\\
hotel-internet: Specifies whether a hotel provides internet access to its guests.\\
train-day: The specific day of the week for a train-related query or reservation .\\
train-destination: The destination for a train.\\
train-departure: The departure location for a train.\\
train-arrive: The desired arrival time at the destination for a train.\\
train-people: The number of people or passengers for a train.\\
train-leave: The desired departure time for a train journey.\\
restaurant-pricerange: The price range of a restaurant.\\
restaurant-day: The specific day of the week for a restaurant.\\
restaurant-time: The time for a restaurant reservation.\\
restaurant-area: The location of a restaurant.\\
restaurant-food: The type or cuisine of food served in a restaurant .
restaurant-people: The number of people or guests for a restaurant-related query or reservation.\\
attraction-area: The location of a attraction.\\
attraction-name: The name or specific name of an attraction.\\
attraction-type: The type or category of an attraction, such as 'museum,' 'park,' 'theater,' etc.
hotel-people: The number of people or guests for a hotel.\\
hotel-area: The location of a hotel\\
hotel-name: The name of a hotel.\\
taxi-leave: The desired departure time for a taxi.\\
taxi-destination: The intended destination for a taxi.\\
taxi-departure: The departure location for a taxi.\\
taxi-arrive: The desired arrival time for a taxi.\\

--------------------\\
\\\noindent
> now make the dialogue state tracking result\\
> The answer must be in JSON format with brace, so that it can be parsed\\
> if there is nothing to inference, the output shuold be not\_mentioned\\
> The answer can be don'tcare\\

prev : [user] i think i would like to visit that location , it seems nice . could i get the phone number and the area ? [system] the area is centre and the phone number is 01223334900 . is there anything else i can help you with ?\\\\
curr : [user] actually , i'm hungry . can you find me a restaurant that serves north american food ? something not super expensive , maybe in a moderate price range ?\\
label:\\}
}
}

\section{Example of Spurious Error}
\label{sec:error}
\normalsize{
In our error analysis, we discovered that the LLM displays intriguing characteristics in relation to the "spurious" type error. We present examples in Table~\ref{tab:error_examples}}
\begin{table*}[h!]

\centering
\scalebox{0.7}{
\begin{tabular}{p{25mm}|p{173mm}}

\hline 
\textbf{Spurious type1}  & \textbf{Predict additional information, by using pre-trained knowledge (Restaurant)} \\\hline
Dial & [sys]  : I would be happy to help. Which restaurant are you looking for?\newline
        [user] : the restaurant is called Eraina. \\
Label & "restaurant-name": "Eraina"\\
Pred & "restaurant-name": "Eraina", \textcolor{blue}{"restaurant-food": "European", "restaurant-pricerange": "expensive"}\\\hline
Dial & [sys]  : What type of food would you like ?\newline
        [user] : Mediterranean food , please  \\
Label & "restaurant-food": "mediterranean", "restaurant-area": "centre"\\
Pred & "restaurant-food": "mediterranean", "restaurant-area": "centre", \textcolor{blue}{"restaurant-pricerange": "expensive"} \\\hline
\textbf{Spurious type2} & \textbf{Predict additional information, by using pre-trained knowledge (Attraction)} \\\hline
Dial &  [user] : May i also get the address for Wandlebury country park please \\
Label &  "attraction-name": "wandlebury country park"\\
Pred & "attraction-name": "wandlebury country park", \textcolor{blue}{"attraction-area": "south", "attraction-type": "park"}\\\hline
Dial &  [user] : How far away is it from the all saints church ?"\\
Label & "None" \\
Pred & \textcolor{blue}{"attraction-type": "architectural", "attraction-area": "centre", "attraction-name": "all saints church"}\\\hline
\textbf{Spurious type3}  & \textbf{Over recommending, by guessing user preference.} \\\hline
Dial & [user] :I'm looking for things to do in the west  \\
Label & "None"\\
Pred & \textcolor{blue}{"restaurant-area": "west", "restaurant-pricerange": "cheap"}\\\hline

\hline

\end{tabular}
}
\caption{Example of spurious error.}
\label{tab:error_examples}

\end{table*}

\end{document}